\documentclass[letterpaper, 10 pt, conference]{ieeeconf}  %

\IEEEoverridecommandlockouts                              %

\usepackage[pdftex]{graphicx}
\usepackage{subcaption}
\usepackage{amsmath} %
\usepackage{amssymb}  %
\usepackage[bookmarks=true]{hyperref}
\usepackage[usenames,dvipsnames,table]{xcolor}
\usepackage{booktabs}
\usepackage{float}
\usepackage{xspace}
\usepackage{tablefootnote}
\usepackage{amsfonts}
\usepackage{multirow}
\usepackage{multicol}
\usepackage{tabularx}
\usepackage[font=small]{caption}
\usepackage{diagbox}
\usepackage{algpseudocode}
\usepackage[ruled,vlined,linesnumbered]{algorithm2e}
\usepackage{makecell}
\usepackage{mathtools}
\usepackage{xcolor}
\usepackage{url}
\usepackage{hyperref}
\hypersetup{
    colorlinks=true,
}

\usepackage{pifont}
\usepackage{wrapfig}
\newcommand{\xmark}{\text{\ding{55}}}

\usepackage{expl3}
\ExplSyntaxOn
\newcommand\latinabbrev[1]{
  \peek_meaning:NTF . {%
    #1\@}%
  { \peek_catcode:NTF a {%
      #1.\@ }%
    {#1.\@}}}
\ExplSyntaxOff
\def\eg{\latinabbrev{e.g}}

\def\ie{\latinabbrev{i.e}}

\makeatletter
    \let\NAT@parse\undefined
\makeatother
\usepackage[square,numbers,sort&compress]{natbib}

\usepackage{hyperref}
\hypersetup{
    colorlinks=true,
    linkcolor=MidnightBlue,
    filecolor=magenta,      
    urlcolor=MidnightBlue,
    citecolor=MidnightBlue,
}
\usepackage[capitalise, nameinlink]{cleveref}

\newcommand\blfootnote[1]{%
  \begingroup
  \renewcommand\thefootnote{}\footnote{#1}%
  \addtocounter{footnote}{-1}%
  \endgroup
}

\renewcommand{\baselinestretch}{0.97}

\def \MethodName {Generalizable Planning-Guided Diffusion Policy Learning}
\def \MethodAcronym {GLIDE}

\graphicspath{{figures/}}

\title{\LARGE \bf
Planning-Guided Diffusion Policy Learning for \\ Generalizable Contact-Rich Bimanual Manipulation
\vspace{-0.5em}
}
\author{Xuanlin Li$^{1,2}$, Tong Zhao$^{1}$, Xinghao Zhu$^{1}$, Jiuguang Wang$^{1}$, Tao Pang$^{1}$, Kuan Fang$^{1,3}$ \\
\url{https://glide-manip.github.io/}
\vspace{-0.2em} 
}%

\begin{document}

\makeatletter
\let\@oldmaketitle\@maketitle%
\renewcommand{\@maketitle}{\@oldmaketitle%
  \captionsetup{type=figure}
  \includegraphics[width=1.0\linewidth]
    {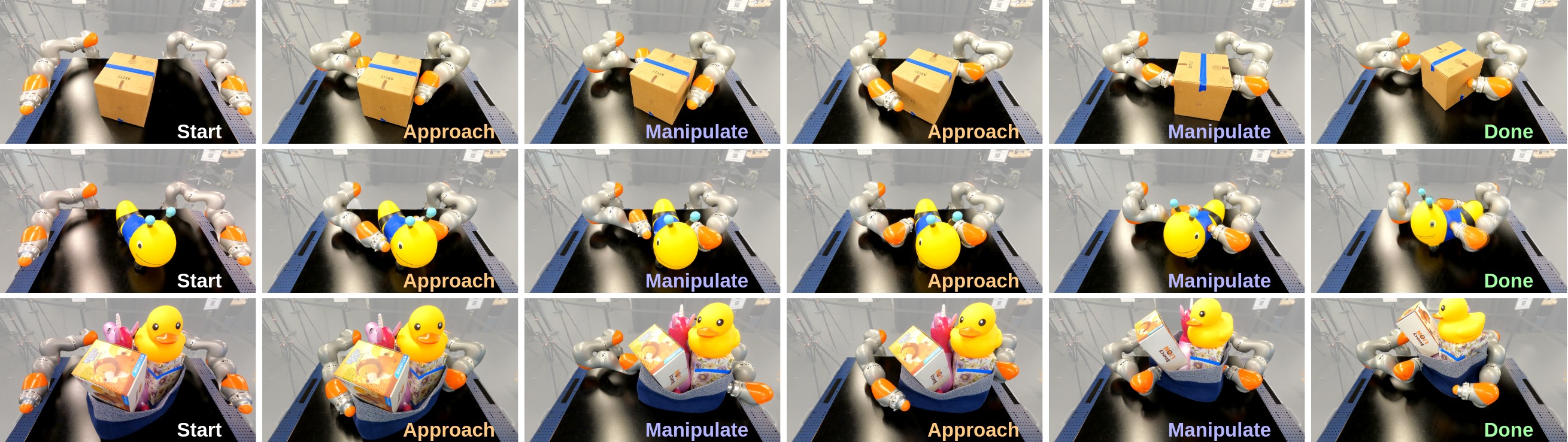}
    \captionof{figure}{\small Contact-rich bimanual manipulation of objects with diverse geometries and physical properties performed by our method. Each row illustrates a trajectory where the robot executes multiple phases of contact (\textit{approach} and \textit{manipulate}) to reorient the object to a target $\mathbb{SE}(2)$ pose. 
    We train a single conditional diffusion policy to control the robot in a closed-loop manner given the observed point clouds.
    }
    \vspace{-0.85em}
    \label{fig:teaser}
    }%
\makeatother

\maketitle

\addtocounter{figure}{-1}
\begin{abstract}
Contact-rich bimanual manipulation involves precise coordination of two arms to change object states through strategically selected contacts and motions.
Due to the inherent complexity of these tasks, acquiring sufficient demonstration data and training policies that generalize to unseen scenarios remain a largely unresolved challenge.
Building on recent advances in planning through contacts, we introduce \MethodName~(\MethodAcronym), 
an approach that effectively learns to solve contact-rich bimanual manipulation tasks by leveraging model-based motion planners to generate demonstration data in high-fidelity physics simulation.
Through efficient planning in randomized environments, our approach generates large-scale and high-quality synthetic motion trajectories for tasks involving diverse objects and transformations. We then train a task-conditioned diffusion policy via behavior cloning using these demonstrations.
To tackle the sim-to-real gap, we propose a set of essential design options in feature extraction, task representation, action prediction, and data augmentation that enable learning robust prediction of smooth action sequences and generalization to unseen scenarios.
Through experiments in both simulation and the real world, we demonstrate that our approach can enable a bimanual robotic system to effectively manipulate objects of diverse geometries, dimensions, and physical properties.
\end{abstract}
\vspace{-1.4em}

\blfootnote{\scriptsize $^{1}$Boston Dynamics AI Institute; $^{2}$UC San Diego; $^{3}$Cornell University
}

\begin{figure*}[t]
    \centering
    \includegraphics[width=1.0\linewidth]{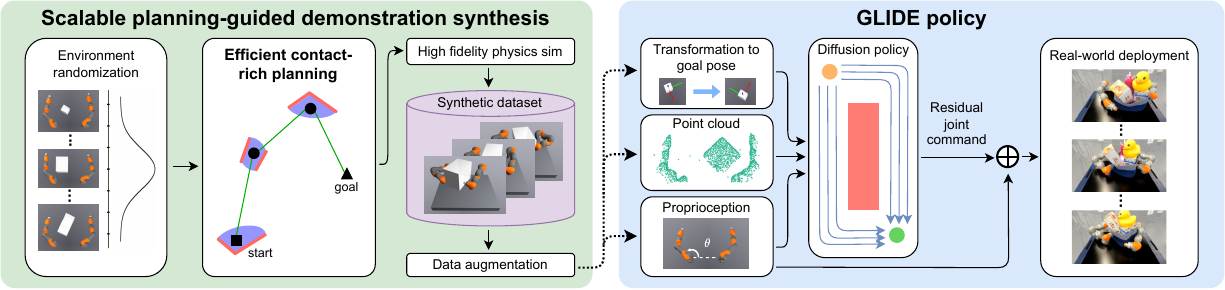}
    \vspace{-1.5em}
    \caption{\small Overview of \textbf{G}eneralizable P\textbf{L}anning-Gu\textbf{I}ded \textbf{D}iffusion Policy L\textbf{E}arning~(\textbf{\MethodAcronym}) for contact-rich bimanual object manipulation. \textbf{Left}: Scalable synthetic demonstration generation using efficient motion planning in physical simulator (Sec.~\ref{sec:method_planning}). \textbf{Right}: Our \MethodAcronym~policy is a task-conditioned point cloud diffusion policy trained through behavior cloning. We further introduce essential design options that significantly enhance our policy's ability to transfer to the real world and generalize to unseen scenarios (Sec.~\ref{sec:method_point_cloud_policy}). 
    }
    \vspace{-1.45em}
    \label{fig:method}
\end{figure*}

\vspace{-0.1cm}
\section{Introduction}
\vspace{-0.1cm}

From warehouse logistics to home services, a broad range of essential robotic applications relies on manipulation involving multi-contact interactions between objects and the manipulators.
For example, as shown in Fig.~\ref{fig:teaser}, the task is to control two robotic arms to manipulate different objects to a specified target pose. 
These objects are often bulky and heavy, making them not directly graspable by the end-effectors. To rearrange and reorient the object, the two arms must hold the object robustly through contacts at multiple links, and then reorient the object over possibly long horizons that take multiple contact phases to reach the goal.
Due to such inherent complexity, solving contact-rich bimanual manipulation for diverse and complex objects remains an open challenge.

To tackle this challenge, recent advances in model-based planning methods utilizing smoothed contact models have begun to demonstrate effectiveness~\citep{tassa2012synthesis, posa2014direct, pang2023global, suh2024dexterous}
However, such motion planners need complete knowledge of object states and environment geometry, thereby limiting their ability to operate in novel environments where objects exhibit diverse geometries and physical properties. Additionally, their computational overhead prevents them from generating trajectories online, which can be a critical limitation, especially in dynamic environments that require real-time adaptation. These limitations highlight the need for approaches that can effectively and robustly perform contact-rich bimanual manipulation over diverse objects.

Alternatively, an increasing number of works have aimed to acquire generalizable contact-rich manipulation skills by learning from trajectory data~\cite{andrychowicz2020learning, chen2023visual, handa2023dextreme, qin2023dexpoint,chi2023diffusion}. In spite of the promising progress, key challenges remain, particularly for bimanual manipulation with complex embodiments.
First, training robust, generalizable policies for complex visuomotor skills usually requires large scale, high-quality trajectory data such as expert demonstrations.
However, it is particularly difficult and costly to collect demonstrations for complex systems and tasks like contact-rich bimanual manipulations using traditional approaches such as teleoperation in the real world~\cite{zhao2023learning, fu2024mobile, chi2024universal}. 
Second, for approaches that learn policies using simulation data, the reality gap in perception and dynamics poses notable challenges for effective policy deployment and generalization. 
Such challenges also become more significant as the complexity of the task increases.

In this work, we propose a method that learns generalizable contact-rich bimanual manipulation, specifically reorientation of bulky and heavy objects, by addressing the aforementioned challenges. Our approach, named \textbf{G}eneralizable P\textbf{L}anning-Gu\textbf{I}ded \textbf{D}iffusion Policy L\textbf{E}arning~(\textbf{\MethodAcronym}), is built upon recent advances in diffusion policies~\cite{chi2023diffusion} and model-based motion planning~\cite{pang2023global} to boost scalability for both data and model.
Using a contact-implicit trajectory optimization solver~\cite{suh2024dexterous}, which derives its efficiency from a smoothed linear approximation of local robot-object contact dynamics, we generate massive and high-quality demonstration data for contact-rich bimanual manipulation in physics simulation.
Compared to prior long-horizon motion planning methods that exhibit high computational costs, this planner significantly improves data generation efficiency with minimal impact on trajectory quality. 
Additionally, we implement our planner to greedily approach the goal object states, further speeding up synthetic data generation.
To learn generalizable visuomotor skills for contact-rich and bimanual manipulation, we build on recent advancements in diffusion policy~\cite{qin2023dexpoint,qi2023general} that effectively capture multimodal action distributions.
In contrast to \cite{qin2023dexpoint,qi2023general}, which learns a specific set of parameters for each task, we design a task-conditioned diffusion policy network that controls the robot to manipulate the object to \textit{arbitrary} target poses specified by the users.
To bridge the reality gap between the simulation and the real-world system, we introduce a set of essential design choices regarding feature extraction, task representation, action prediction, and data augmentation, which significantly improve our policy's performance in sim-to-real transfer. 
We evaluate \MethodAcronym~in a series of simulated and real-world contact-rich bimanual manipulation tasks involving both in-distribution and out-of-distribution (OOD) objects. 
Through detailed analysis, we demonstrate that our approach can robustly accomplish contact-rich bimanual manipulation in unseen scenarios.

\vspace{-0.5em}
\section{Related Work}
\vspace{-0.15em}

\noindent \textbf{Contact-rich and bimanual manipulation} have been widely studied, with key challenges arising from non-smooth contact dynamics and navigating non-convex cost landscapes~\cite{pang2023global}. Previous work in planning has formulated contact-rich manipulation as a Mixed-Integer Program (MIP), using integer variables to represent contact modes~\cite{marcucci2017approximate, hogan2020feedback, aydinoglu2024consensus}. While MIP-based methods offer optimality, they scale poorly for tasks like full-arm bimanual manipulation due to the exponential growth in modes with increasing contact pairs and task horizons. Recent planning approaches have improved efficiency by sacrificing optimality for speed through analytic smoothing~\cite{pang2023global,howell2022dojo,suh2024dexterous}, but planners rely on full knowledge of object states and geometry, limiting their generalization to diverse real-world objects. To address generalization, another line of prior work has approached contact-rich manipulation through behavior cloning and reinforcement learning~\citep{bertsekas2019reinforcement,andrychowicz2020learning, chen2023visual, handa2023dextreme, qin2023dexpoint,grannen2023stabilize}. While they have shown promising results, few have focused on generalizable bimanual contact-rich manipulation due to its inherent complexity. A recent work~\cite{lin2023bitouch} has made progress in this area, but it only uses two end-effectors with single contact points to perform manipulation rather than the full bimanual arms, and it lacks visual feedback. This limits its ability to manipulate bulky or lengthy objects. In this work, we propose a method that demonstrates generalizable contact-rich bimanual manipulation over objects with diverse dimensions, geometries, and physical properties, utilizing visual feedback and without assuming privileged information (e.g., object shapes) to be known.

\noindent \textbf{Visuomotor policy learning for manipulation.} 
Modern machine learning approaches have enabled robots to learn visuomotor policies for a variety of tasks like grasping, pushing, rotating, and insertion~\cite{mandlekar2018roboturk, mandlekar2019scaling, eppner2021acronym, ebert2021bridge, zhu2022learn, liu2022frame, zhang2022learning,Seita2022toolflownet}.
Compared to these primitive skills, contact-rich bimanual manipulation introduces critical challenges to policy learning due to the complex dynamics and multimodal distribution of behaviors.
With recent advances in deep generative models~\cite{ajay2022conditional, chen2021decision}, diffusion policies~\cite{chi2023diffusion, janner2022planning, hansenestruch2023idqlimplicitqlearningactorcritic, reuss2023goalconditionedimitationlearningusing, pearce2023imitatinghumanbehaviourdiffusion} have been introduced as a new class of policy networks that are capable of capturing complex data distributions. 
While these methods have shown promise, they are often applied to specific tasks with fixed goals.
In this work, we design a diffusion policy network for contact-rich bimanual manipulation given point clouds.
To facilitate generalization, we propose a planning-guided learning approach that trains the policy on expert demonstrations with diverse objects generated in simulation.

\noindent \textbf{Planning-guided data synthesis.} While traditional motion planning systems require complete knowledge of scenes and objects, limiting their real-world deployment, they are highly valuable for synthetic data generation for learning-based methods. Prior imitation learning work has used planning data to train effective, generalizable policies for collision avoidance~\cite{fishman2023motion,dalal2024neuralmp}, manipulation~\cite{dalal2023imitating,tosun2019pixels,yamada2021motion}, locomotion~\cite{carius2020mpc,abbatematteo2021bootstrapping}, and autonomous driving~\cite{Pan2017AgileAD}. Model-based planning has also been widely used to improve sample efficiency and performance in reinforcement learning~\cite{jurgenson2019harnessing, ha2020learning, levine2013guided, zhang2023plan, jenelten2024dtc, brudigam2024jacta}. However, using planning to generate high-quality demonstrations for contact-rich bimanual manipulation scalably remains underexplored due to the challenges posed by complex contact dynamics and planning efficiency. In this work, we build on recent advances in trajectory optimization and motion planning~\cite{pang2023global,suh2024dexterous} to effectively and scalably generate demonstrations for contact-rich bimanual manipulation.

\vspace{-0.3em}
\section{Method}
\vspace{-0.1em}
{\MethodAcronym} centers around two key focuses: generating diverse and high-quality training data for contact-rich bimanual manipulation tasks and learning visuomotor policies that can generalize to unseen environments and task specifications.

In this section, we will first describe the problem formulation for learning generalizable contact-rich bimanual manipulation. 
Next, we will propose a data synthesis pipeline that uses an efficient contact-rich planner.
Finally, we will devise a conditional diffusion policy that generates action sequences given observed point clouds and task specifications.

\vspace{-0.4em}
\subsection{Problem Formulation}
\vspace{-0.2em}
\label{sec:problem_formulation}

We consider the problem of controlling a bimanual robotic system to change the pose of objects. Such objects can be bulky and heavy, and thus cannot be directly grasped by the end-effectors.
To accomplish this task, the two robot arms need to strategically approach the object to make contact and then reorient the object to the target pose. 
For complex objects and challenging target poses, this process might take multiple rounds of approaching and manipulation due to the limitation of the configuration space of the robot.
To enable generalization to unseen objects, we do not assume the shape and initial pose of the object to be known.
Instead, the robot only receives visual observations of the environment.

Formally, we define the problem with the environment state space $\mathcal{S}$, the observation space $\mathcal{O}$, the action space $\mathcal{A}$, the task space $\mathcal{C}$, and the time horizon $H$.
In each episode, the robot starts at an initial environment state $s_0 \in \mathcal{S}$ and is commanded to change the environment according to a task specification $c \in \mathcal{C}$. At each time step $t = 1,...,H$, the robot receives the observation $o_t \in \mathcal{O}$ and takes an action $a_t \in \mathcal{A}$.
To solve this problem, we aim to learn a single policy $\pi_{\theta}(a | o, c)$ with a trainable set of parameters $\theta$. 
Due to the complexity of the dynamics and control in bimanual manipulation tasks, directly collecting massive and high-quality data in the real world is hard. Instead, we aim to generate \textit{synthetic} demonstration data $\mathcal{D}$ to train the policy $\pi$.
As no real-world training data is collected and used, we would need to design the policy network and learning algorithm such that the policy trained on $\mathcal{D}$ is transferrable to the real world at deployment.

In this work, we specifically consider a table-top environment involving two 7-DoF robotic arms without end-effectors, as shown in Fig.~\ref{fig:teaser}.
The policy does not know the true environment state $s_t$ but receives the observation $o_t$ as a depth image taken from a fixed RGBD camera converted into point clouds as well as the proprioceptive joint states of the robot.
The target object to manipulate is supported by the table surface, and the task is defined by a transformation of the object pose in $\mathbb{SE}(2)$.
At each time step, we compute the task specification $c_t$ as the transformation between the current object pose and the target pose as part of the inputs to the policy $\pi$.

\vspace{-0.4em}
\subsection{Demonstration Synthesis via Efficient Planning}
\vspace{-0.2em}
\label{sec:method_planning}

To scalably generate demonstrations for our bimanual object reorientation task, we use motion planning that leverages privileged object and robot states in simulation to produce trajectories.
We build upon the planning-through-contact framework in \cite{pang2023global} and incorporate the very recent advances in \cite{suh2024dexterous} in our planning pipeline, which proposes a smoothed linear approximation of local robot-object contact dynamics that significantly improves planning efficiency. 

Our planner's input include (\textbf{i}) the initial state of the system, $s_0 \coloneq (q_0^a, q_0^u)$, where $q_0^a$ is the robot joint angles and $q_0^u$ the object pose; and (\textbf{ii}) the goal object pose $q_\mathrm{goal}^u$. An action $a$ consists of commanded joint angles for both arms. The planner generates a sequence of actions $T \coloneq (a_0, a_1, \dots)$ that takes the object from the initial pose to the goal pose. Compared with with the sampling-based planner in \cite{pang2023global}, we greedily approach the goal to speed up computation and encourage consistency in the demonstrations. The greedy approach did not significantly impact the planner's success rate in practice.
The resulting planner, summarized in \cref{alg:plan}, consists of the following components:
\begin{itemize} %
    \item A \textit{contact sampler} that generates robot joint configurations $q^a_\text{grasp}$ where the robot arms make contact with the object in a way that facilitates manipulation (\ie, a \textit{``grasp''}). In our implementation, we use inverse kinematics to generate grasps where the robot's distal links can stably pinch and hold the object.
    \item A \textit{collision-free planner} using bidirectional RRT~\citep{lavalle2006planning} with shortcutting to plan a collision-free trajectory from the current robot joint configuration $q^a$ to the next grasp $q^a_\text{grasp}$.
    \item A \textit{contact planner} which, given a current robot configuration $q^a$ that is already grasping the object and the object's current configuration $q^u$, greedily moves the object towards the goal configuration $q^u_\text{goal}$ as much as possible while ensuring that the robot does not exceed joint limits.
\end{itemize}

\vspace{-1.2em}
\begin{algorithm}
\small
\caption{\small{\textbf{Demonstration Synthesis via Planning}}}\label{alg:plan}
\textbf{Input:} $q^a_0, q^u_0, q^u_{\mathrm{goal}}$ \\
\textbf{Output:} Action trajectory $T$ \\
$q^a , \ q^u , \ T \gets q^a_0 , \ q^u_0$ , \texttt{list()}\\
\While{$q^u \neq q^u_{\mathrm{goal}}$}{
    $q^a_{\text{grasp}} \gets \textsc{SampleContact}(q^u)$ \\
    \While{$q^a \neq q^a_{\text{grasp}}$}{
        $q^a, a \gets \textsc{PlanCollisionFree}(q^a,q^a_\mathrm{grasp})$ \\
        $T.$\texttt{extend}$(a)$ \\
    }
    \While{$q^a$ not at joint limit and $q^u \neq q^u_\mathrm{goal}$}{
        $q^a, q^u, a \gets \textsc{PlanContact}(q^a,q^u, q^u_\mathrm{goal})$ \\
        $T.$\texttt{extend}$(a)$ \\
    }
}
\algorithmicreturn \ $T$
\end{algorithm}
\vspace{-1.2em}

\noindent \textbf{Contact Planner Details.} 
Much prior work has focused on trajectory optimization through contact \cite{posa2014direct,aydinoglu2024consensus,lecleach2024fast,suh2022bundled,tassa2012synthesis}, but their high computational costs due to long-horizon trajectory optimization and the exponential number of contact modes prohibit efficient trajectory generation. To reduce these costs, we adopt a \textit{single step} variant of trajectory optimization and solve it through the approach in~\citep{suh2024dexterous}. In short, we use a linear approximation $f_{\text{local}}$ of the local contact dynamics to solve the optimization problem with the following objective:
{
\setlength{\abovedisplayskip}{0.8pt}
\setlength{\belowdisplayskip}{0.8pt}
\[
    \min_{q^u_+, a} \ (q^u_+ - q^u_\text{goal})^T{\mathbf{Q}}(q^u_+ - q^u_\text{goal}) + (a - q^a)^T\mathbf{R}(a - q^a)
\]
}
Here, $q^u_+ = f_{\text{local}}(q^u, q^a, a)$ denotes the object's approximate configuration after the robot action $a$ is taken, and $\mathbf{Q}$ and $\mathbf{R}$ are user-specified cost matrices. 

\noindent \textbf{Filtered Behavior Cloning.} 
As the collected trajectories are often suboptimal due to the approximality of the contact dynamics and the stochasticity of the RRT in the planner, directly performing behavior cloning on all trajectories leads to poor policy performance. To address this, we extract high-quality demonstrations by filtering trajectories via rollouts in a high-fidelity simulator~\cite{drake} to verify their accuracy. The trajectories in which the object doesn't reach the goal and the suboptimal trajectories that take too long to reach the goal are discarded in this process. 
Finally, we rebalance the trajectories to be uniformly distributed across objects and render uncolored point clouds for policy training.
We save the resultant subset of demonstration trajectories as the dataset $\mathcal{D}$ to train the policy.

\vspace{-0.4em}
\subsection{Diffusion Policy Learning from Synthetic Demonstrations}
\vspace{-0.2em}
\label{sec:method_point_cloud_policy}

Given the synthetic demonstrations $\mathcal{D}$, we train a diffusion policy $\pi_\theta(a | s, c)$ for contact-rich bimanual manipulation via behavior cloning.
To enable $\pi_\theta$ to predict suitable manipulation actions for unseen scenarios and effectively transfer the learned knowledge to the real world, we introduce a set of essential design options to existing diffusion policy learning methods~\citep{chi2023diffusion} for the feature extraction, task representation, and action prediction.

To extract the geometric information of the environment from the noisy point cloud observations, we design a feature extraction backbone for $\pi_\theta$ that can facilitate generalization to unseen objects based on \citep{chi2023diffusion} and \citep{Ze2024DP3}.
To enable generalization to unseen environments, we clip the point clouds within the robot's workspace and remove the irrelevant background objects in each frame. 
Additionally, to address the reality gap and sensor noise encountered in the real world, we introduce a \textit{Flying Point Augmentation} approach, where we randomly add large Gaussian noise to the points with a small probability (\eg, 0.5\%). We find that this approach significantly improves our policy's real-world performance while requiring minimal implementation effort.

While \citep{chi2023diffusion} and \citep{Ze2024DP3} aim to learn a predefined set of tasks using a different set of parameters for each task, we train a \textit{single} policy that takes the task specification $c$ as an additional input. 
As described in Sec.~\ref{sec:problem_formulation}, instead of assuming the object shape is known and directly specifying a target object pose, we implicitly specify the target pose using the initial visual observation $o_0$ along with the delta transformation $c_0$ from the initial object pose in $o_0$ to the target object pose. 
In each subsequent time step $t$, we recompute the transformation $c_t$ from the current pose to the target pose given the current observation $o_t$.
Without knowing the object shape and pre-defining an object frame, we propose to obtain $c_t$ by segmenting the target object in $o_0$ using an open-vocabulary segmentation algorithm~\citep{liu2023grounding,cai2023efficientvit}, selecting keypoints using the farthest point sampling within the segmentation, and then tracking the keypoints in the 3D space through real-time object tracking~\cite{doersch2023tapir,karaev2023cotracker}.

We design the prediction head of the policy network to robustly generate smooth and feasible motion trajectories.
Following~\cite{chi2023diffusion}, our policy predicts an action sequence of $T_a$ steps. We use a larger $T_a = 20$ at test time to improve performance and train our policy with $T_a = 64$.
Additionally, prior work like~\citep{chi2023diffusion} usually directly predicts the absolute end-effector poses or joint angles as actions.
However, we observe that this can result in poor generalization to unseen objects and non-smooth trajectories in the real world.
To address this issue, we instead re-design the prediction head to predict the \textit{residual} joint position actions $a_{t+1:t+T_a} = \{q_i - q_{t}\}_{i=t+1}^{t+T_a}$ with $q_{t}$ being the current joint positions at time step $t$.
Compared to absolute joint actions, the residual joint actions are much more consistent in scales and shifts across training trajectories, resulting in significantly better real-world policy performance.

\begin{figure}[t]
    \centering
    \includegraphics[width=0.95\linewidth]{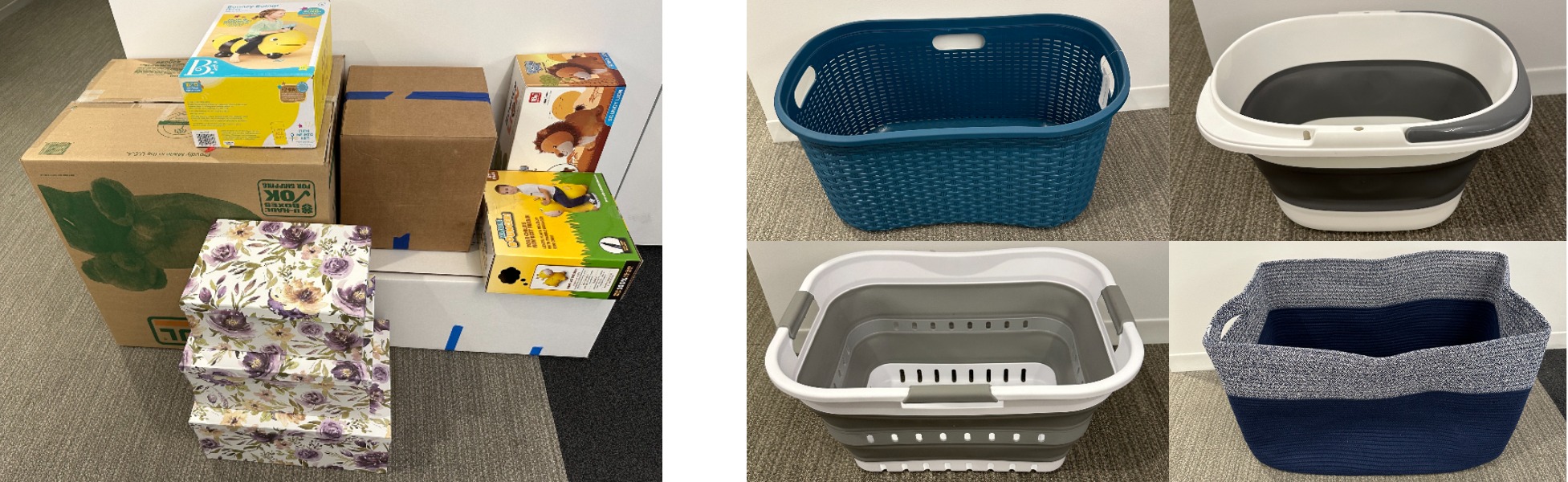}
    \caption{
        Objects used for real-world evaluation: \textbf{(Left)} Boxes whose dimensions and physical properties are within our policy's training distribution; \textbf{(Right)} Out-of-distribution (OOD) containers, some of which are made with soft materials like rubber (top right, bottom left) and fabric (bottom right). For half of OOD evaluations, we evaluate our policy using empty containers; for the other half, we evaluate using containers overfilled with miscellaneous objects.
    }
    \vspace{0.5em}
    \label{fig:real_obj_vis}
\end{figure}

\vspace{-0.3em}
\section{Experiments}
\vspace{-0.15em}

We design our experiments to answer the following questions: \textbf{(1)} Does \MethodAcronym~effectively perform contact-rich bimanual object manipulation on objects whose geometries, dimensions, and physical properties are within the distribution of training demonstrations? \textbf{(2)} How well does \MethodAcronym~generalize to objects with challenging geometries and physical properties? \textbf{(3)} How important are the design choices in \MethodAcronym~to the performance? 
\vspace{-0.4em}
\subsection{Environments and Tasks}
\vspace{-0.2em}
\label{sec:experimental_setup}

We perform bimanual object reorientation on a flat table using two 7-DoF KUKA LBR iiwa arms in both the \textsc{Drake}~\citep{drake} simulator and the real world. We use a Realsense D455 camera mounted 2 meters above the table to capture real-world visual observations, and we actuate three joints per arm following~\citep{zhu2024learning}. In simulation, we synthesize planner demonstrations following our approach in Sec.~\ref{sec:method_planning} and then retain 12,000 successful trajectories for policy training (Sec.~\ref{sec:method_point_cloud_policy}).  During both demonstration generation and policy evaluation, we randomize object assets along with their initial and goal poses. An episode is successful if the final object pose is within 10 cm and 0.2 rad of the goal. 
The demonstration generation process takes about two days on a 96-CPU machine.

For object assets used for demonstration generation and policy training, we generate 2,000 rectangular box primitives with randomized dimensions, mass, and friction coefficients. While we train our policy exclusively on box primitives, we show that it can achieve good generalization performance on objects with OOD geometries and physical properties.

\vspace{-0.5em}
\subsection{Experiment Design}
\vspace{-0.2em}
\label{sec:eval_protocol}

\begin{table}
        \centering
        \scriptsize
        \setlength{\tabcolsep}{3pt}
        \renewcommand{\arraystretch}{0.9}
        \begin{tabular}{@{}lccc@{}}
        \toprule
        Task (In-Dist Eval) & Planner (Sim) & Policy (Sim) & Policy (Real) \\
        \midrule
        Fixed 45$^\circ$ Rotation  & 0.337 & 0.740 & 0.800 \\
        \midrule
        Random Rotation (Easy) & 0.227  & 0.610 & 0.600 \\
        Random Rotation (Medium) & 0.141 & 0.410  & 0.360 \\
        Random Rotation (Hard) & 0.099 & 0.180 & 0.200 \\
        Random Rotation (Overall) & 0.156 & 0.400 & 0.387 \\
        \bottomrule
        \end{tabular}
        \vspace{-0.15em}
        \caption{Bimanual object reorientation success rates in both simulation and the real world for in-distribution objects. We report results for our point cloud policy and also include the planner results as a reference (note that the planner runs significantly slower than our policy and requires access to complete knowledge of object geometry, and is thus infeasible for real world deployment). For vanilla DP3 baseline results, see the last row of Tab.~\ref{tab:ablation_sim2real} and the first column of Tab.~\ref{tab:ablation_n_action}.}         
        \label{tab:result_in_distribution}
        \vspace{0.5em}
        \renewcommand{\arraystretch}{0.8}
        \begin{tabular}{@{}lccc@{}}
        \toprule
        Task (Real OOD Eval) & Empty Containers & Overfilled Containers & Overall \\
        \midrule
        Fixed 45$^\circ$ Rotation  & 0.688 & 0.625 & 0.657 \\
        \midrule
        Random Rotation & 0.250 & 0.313  &  0.282 \\
        \bottomrule
        \end{tabular}
        \vspace{-0.15em}
        \caption{Success rates of real-world policy evaluation using out-of-distribution containers (which are visualized in Fig.~\ref{fig:real_obj_vis} - Right). We evaluate our policy using both empty and overfilled containers.}
        \label{tab:result_ood}
        \vspace{0.5em}
\end{table}

\begin{figure}[t]
    \centering
    \includegraphics[width=0.95\linewidth]{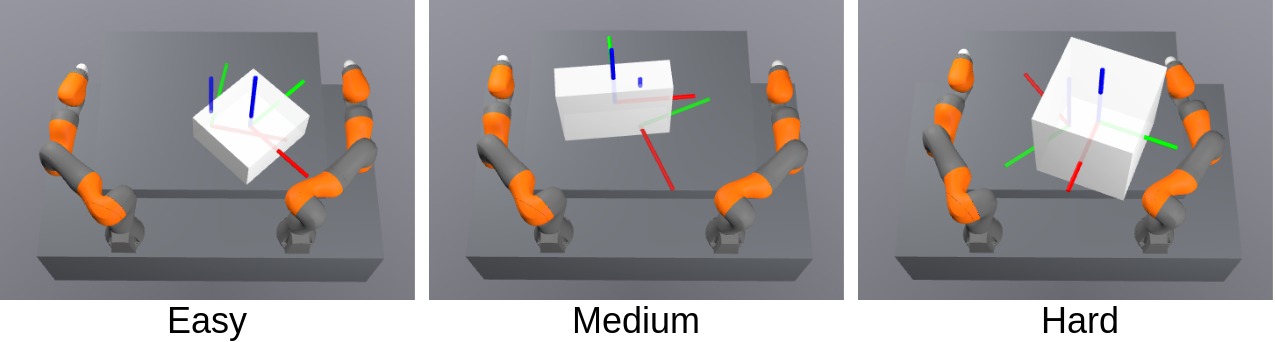}
    \vspace{-0.4em}
    \caption{
        Visualization of our bimanual box reorientation environment in the Drake simulator. The coordinate frame on the box represents its current pose, while the coordinate frame on the table indicates the goal pose. We illustrate 3 environment variations with randomized boxes and increasing difficulty levels: easy ($|\Delta \theta| \le 45^{\circ}$), medium ($45 < |\Delta \theta| \le 90^{\circ}$), and hard ($90^{\circ} < |\Delta \theta| \le 150^{\circ}$). 
    }
    \label{fig:sim_env_vis}
    \vspace{0.5em}
    \includegraphics[width=0.95\linewidth]{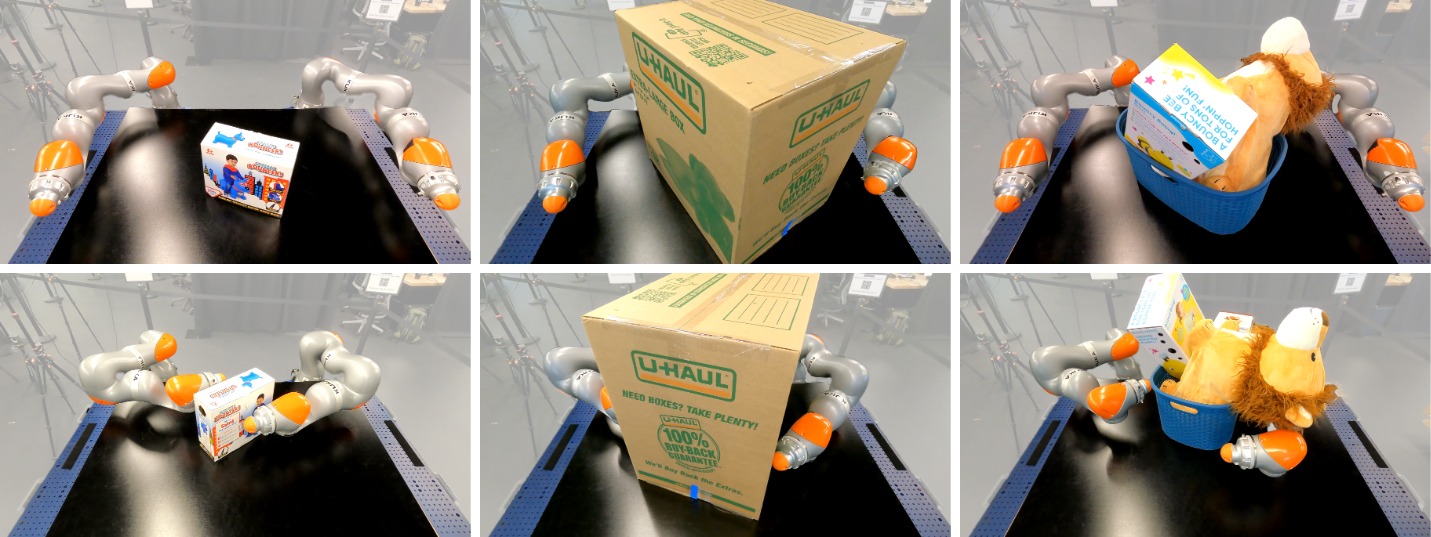}
    \vspace{-0.25em}
    \caption{
        Examples of policy evaluation on in-distribution (left, middle) and out-of-distribution (right) objects. The first row shows the initial state, and the second row shows the final state.
    }
    \label{fig:id_ood_eval}
    \vspace{0.5em}
    \includegraphics[width=0.95\linewidth]{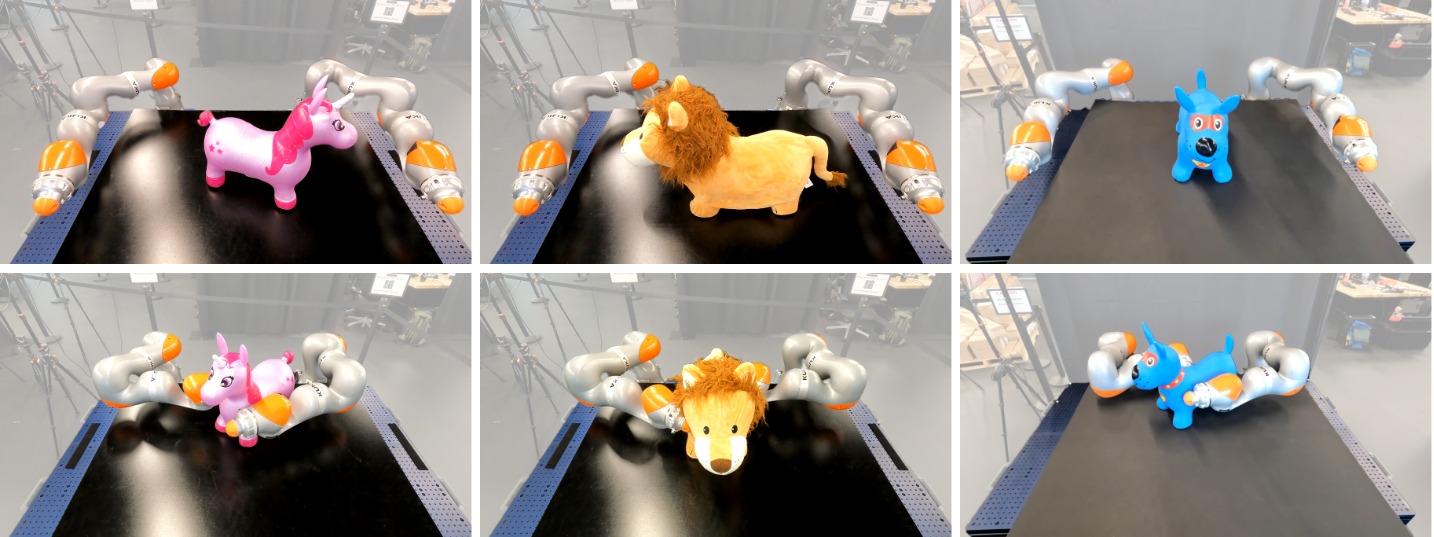}
    \vspace{-0.25em}
    \caption{
        Visualization of policy evaluation under a challenging OOD scenario using irregularly shaped toys. The success rate is 52\% on the fixed 45$^{\circ}$ rotation task and 28\% on the random rotation task.
    }
    \vspace{0.5em}
    \label{fig:ood_toy_eval}
\end{figure}

To comprehensively assess \MethodAcronym's performance in contact-rich bimanual object reorientation, we evaluate it on objects with geometries, dimensions, and physical properties within the distribution of demonstration trajectories (\ie, \textit{``In-Distribution Evaluation''}). Additionally, we put \MethodAcronym~to the test and evaluate its ability to generalize to challenging, unseen objects outside of training distribution (\textit{``Out-of-Distribution (OOD) Evaluation''}). Despite the perception and control challenges posed by the diverse shapes and physical properties of OOD real world objects, we hope that our policy can still effectively infer object boundaries from our processed point clouds where irrelevant backgrounds are removed, from which it can determine the necessary robot joint movements to approach and manipulate the object.

Furthermore, to enable a fine-grained analysis of \MethodAcronym's success and limitations, we design our experiments to encompass increasing levels of task difficulty:
{
\setlength{\abovedisplayskip}{0.2pt}
\setlength{\belowdisplayskip}{0.2pt}
\begin{itemize} %
    \item \textbf{Fixed Rotation}: Let $\Delta \theta$ represent the orientation difference between the initial and goal object poses. In our "Fixed Rotation" setting, we set $\Delta \theta$ to a constant value across all demonstration and evaluation episodes (in our experiments, $\Delta \theta = 45^{\circ}$). Additionally, we initialize object orientations in multiples of $\frac{\pi}{2}$ (for boxes, their edges are aligned parallel to the table edges). Although simpler than the ``Random Rotation'' setting below, this setup is already challenging as it requires the policy to generalize to diverse objects with randomized positions.
    \item \textbf{Random Rotation}: Both $\Delta \theta$ and the initial object orientation are fully randomized. We further subdivide $\Delta \theta$ into 3 levels of difficulties (visualized in Fig.~\ref{fig:sim_env_vis}): \textit{Easy} ($|\Delta \theta| \le 45^{\circ}$); \textit{Medium} ($45 < |\Delta \theta| \le 90^{\circ}$); \textit{Hard} ($90^{\circ} < |\Delta \theta| \le 150^{\circ}$)\footnote{We set $|\Delta \theta| \le 150^{\circ}$ as it is very challenging to generate successful demonstrations for $|\Delta \theta| > 150^{\circ}$ with our current planner.}. As $|\Delta \theta|$ increases, it becomes more likely that the robot needs to perform multiple rounds of approaching and manipulation to reach the object to the goal orientation due to the limitation of robot's configuration space.
\end{itemize}
}
We generate demonstration sets for both the ``Fixed Rotation'' and ``Random Rotation'' tasks and train a diffusion policy on each. For the ``Random Rotation'' task, we evaluate a single policy across all three levels of task difficulty.

\vspace{-0.4em}
\subsection{In-Distribution Evaluation}
\vspace{-0.2em}
\label{sec:exp_id_evaluation}

We perform in-distribution evaluation of \MethodAcronym~in both simulation and the real world. In simulation, we randomly sample boxes whose dimensions and physical properties are within our demonstration's distribution. In the real world, we randomly choose a set of everyday boxes from the same distribution, as shown in Fig.~\ref{fig:real_obj_vis}. We conduct 100 policy evaluation trials in simulation and 25 trials in the real world. 
Additionally, we include our motion planner's success rates in simulation as a reference.
Note that the planner runs slower than our policy and requires access to complete knowledge of object geometry, which is infeasible for real-world evaluation. Due to the highly stochastic nature of our planner, we evaluate it over a larger number of 600 trials.

We present our results in Tab.~\ref{tab:result_in_distribution}, and we include qualitative visualizations of policy rollouts in the first row of Fig.~\ref{fig:teaser} and the left and middle columns of Fig.~\ref{fig:id_ood_eval}. Overall, \MethodAcronym~demonstrates strong performance on in-distribution objects in both simulation and the real world. Notably, by guiding diffusion policy learning through motion planning, our policy significantly improves the performance over motion planning while being much faster and does not require prior knowledge of object shapes. However, there is still significant room for improvement in the highly challenging setting where the policy must manipulate the object towards a distant target orientation over a long horizon. Later in Sec.~\ref{sec:failure_mode}, we will analyze these failure cases in detail and suggest directions for future improvements.

\vspace{-0.4em}
\subsection{Out-of-Distribution (OOD) Evaluation}
\vspace{-0.2em}
\label{sec:exp_ood_evaluation}

We evaluate \MethodAcronym's ability to generalize to out-of-distribution objects in the real world using a set of containers shown in Fig.~\ref{fig:real_obj_vis}-Right. These containers differ from the rigid boxes in both geometry and physical properties: some have curved edges or a wider top than the bottom, while others are deformable, made from soft materials like rubber or fabric. We conduct 16 evaluation trials with the containers empty and another 16 trials with them \textit{overfilled} with miscellaneous objects. As illustrated in the third row of Fig.~\ref{fig:teaser} and the right column of Fig.~\ref{fig:id_ood_eval}, overfilled containers introduce irregular object boundaries and object weights heavier than those during training, adding further challenges for our policy.
Results are shown in Tab.~\ref{tab:result_ood}. \MethodAcronym~demonstrates robustness and generalization to OOD shapes and object properties, with a slight performance drop compared to in-distribution evaluation in Tab.~\ref{tab:result_in_distribution}. Notably, the success rates are similar across both empty and overfilled containers, despite their distinct geometries and weight, showcasing \MethodAcronym's robustness in handling diverse and challenging scenarios.

Additionally, we evaluate \MethodAcronym~on inflatable toys as shown in Fig.~\ref{fig:ood_toy_eval} and the second row of Fig.~\ref{fig:teaser}. These toys present even larger distribution shifts from the training objects, featuring legged bases instead of the rectangular bottoms seen in boxes and containers. Additionally, their higher center of mass makes them prone to tipping if the two robot arms do not act with precise coordination. \MethodAcronym~achieves 52\% for the fixed rotation task and 28\% for the random rotation task, further demonstrating its potential to adapt and generalize to challenging objects.

\vspace{-0.4em}
\subsection{Ablations}
\vspace{-0.2em}

In this section, we analyze the importance of different design choices to \MethodAcronym's performance. In Tab.~\ref{tab:ablation_sim2real}, we ablate the residual robot joint action prediction and the flying point augmentation approaches we introduced in Sec.~\ref{sec:method_point_cloud_policy} to improve \MethodAcronym's sim-to-real transfer. While these techniques show minimal impact in simulation, they significantly enhance policy performance in real-world evaluations. Omitting either approach results in a notable drop in real-world performance. In Tab.~\ref{tab:ablation_n_action}, we ablate the number of action prediction steps $T_a$ for our diffusion policy evaluation. We find that a larger $T_a=20$, compared to prior work~\citep{chi2023diffusion} with $T_a = 8$, results in smoother trajectories and better performance. However, further increasing $T_a$ can degrade performance due to the lack of real-time feedback, which limits the policy's ability to make timely adaptations during object manipulation. In Tab.~\ref{tab:ablation_n_demo}, we study how the number of planner-generated demonstrations impacts policy performance. We find that thousands of demonstrations are needed for good policy performance on the easier ``Fixed Rotation'' task, with significantly more required for the challenging ``Random Rotation'' task. We also find that the performance doesn't plateau as we scale up the number of demonstrations, suggesting that further scaling up demonstration generation will continue to improve policy effectiveness.

\begin{table}[t]
\vspace{1.0em}
        \centering
        \scriptsize
        \setlength{\tabcolsep}{2.2pt}
        \renewcommand{\arraystretch}{0.9}
        \begin{tabular}{@{}cc|cc@{}}
        \toprule
        Residual Action & Flying Point Aug & Success (Sim, In-Dist) & Success (Real, In-Dist) \\
        \midrule
        $\checkmark$ & $\checkmark$ & 0.740 & 0.800 \\
        $\checkmark$ & $\xmark$ & 0.750 & 0.320 \\
        $\xmark$ & $\checkmark$ & 0.780 & 0.520 \\
        $\xmark$ & $\xmark$ & 0.750 & 0.000 \\
        \bottomrule
        \end{tabular}
        \vspace{-0.15em}
        \caption{Ablation of different point cloud diffusion policy design choices on our fixed 45$^\circ$ clockwise object rotation task.} 
        \label{tab:ablation_sim2real}
        \vspace{0.5em}
        \setlength{\tabcolsep}{4.8pt}
        \renewcommand{\arraystretch}{0.6}
        \begin{tabular}{@{}lcccc@{}}
        \toprule
        Task (Sim, In-Dist Eval) & $T_a=8$ & $T_a=20$ & $T_a=40$ & $T_a=64$ \\
        \midrule
        Fixed 45$^\circ$ Rotation  & 0.440 & 0.740 & 0.760  & 0.770 \\
        \midrule
        Random Rotation & 0.270 & 0.400 & 0.340 & 0.200 \\
        \bottomrule
        \end{tabular}
        \vspace{-0.15em}
        \caption{Ablation on the number of action prediction steps $T_a$ for diffusion policy evaluation. We use $T_a=64$ for training.}
        \label{tab:ablation_n_action}
        \vspace{0.5em}
        \setlength{\tabcolsep}{3.0pt}
        \renewcommand{\arraystretch}{0.6}
        \begin{tabular}{@{}lcccc@{}}
        \toprule
        Task (Sim, In-Dist Eval) & 500 Demos & 2500 Demos & 7500 Demos & 12000 Demos \\
        \midrule
        Fixed 45$^\circ$ Rotation  & 0.330 & 0.690 & 0.740  & 0.800 \\
        \midrule
        Random Rotation & 0.030 & 0.170 & 0.330 & 0.400 \\
        \bottomrule
        \end{tabular}
        \vspace{-0.15em}
        \caption{Ablation on the number of expert planner demonstrations for diffusion policy learning.}
        \vspace{0.5em}
        \label{tab:ablation_n_demo}
\end{table}

\vspace{-0.45em}
\subsection{Failure Mode Analysis}
\vspace{-0.25em}
\label{sec:failure_mode}

While our method demonstrates effective generalizable bimanual contact-rich manipulation, there is still room for improvement on highly challenging scenarios requiring manipulating objects to distant target orientations.
\begin{wrapfigure}[6]{r}{0.20\textwidth}
    \vspace{-1.2em}
    \centering
    \includegraphics[width=\linewidth]{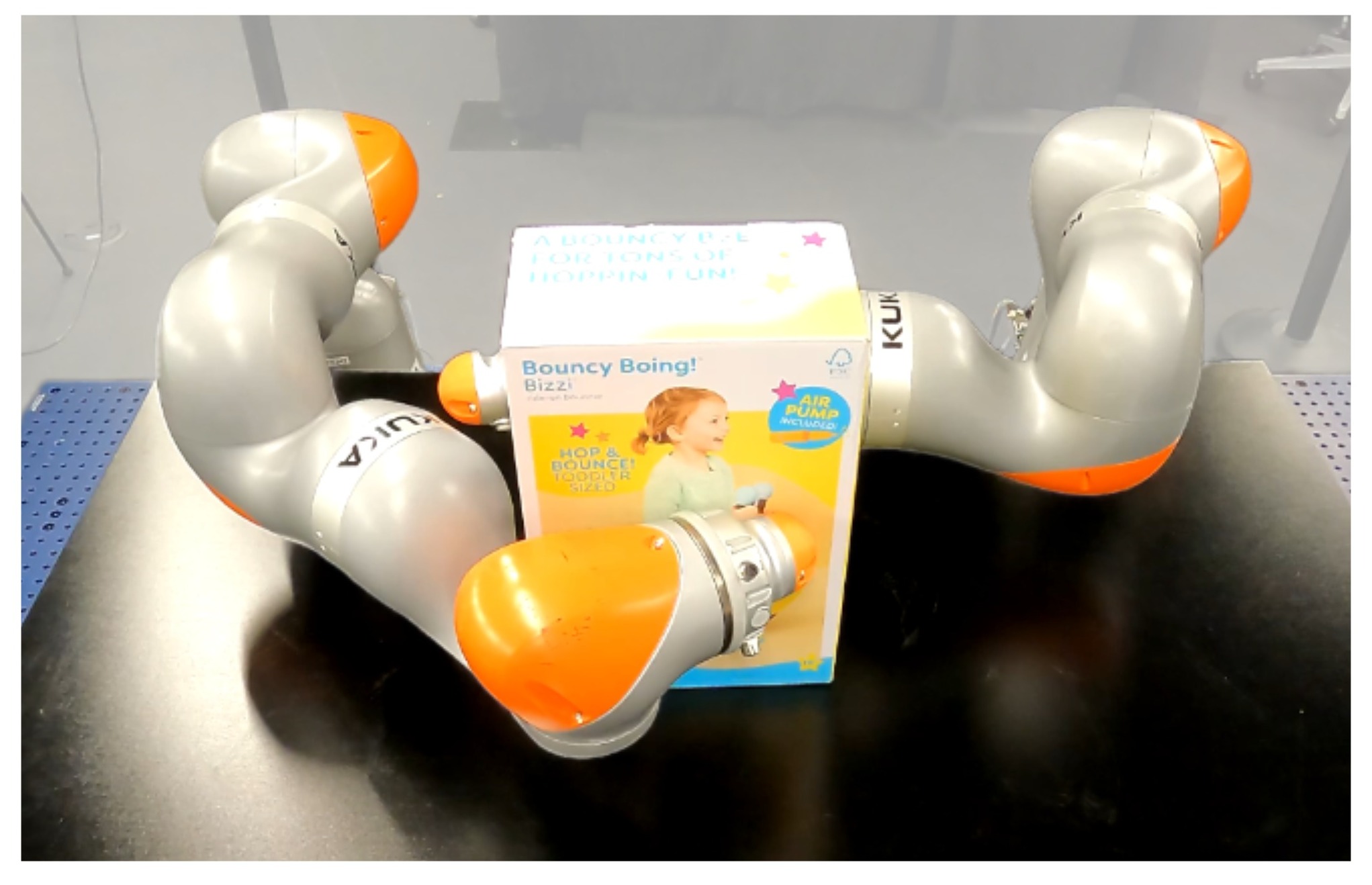}
\end{wrapfigure}
Among 52\% of failure cases in our ``Random Rotation-Hard'' experiments, the robot is stuck in a poor joint configuration that prevents further object rotation, as illustrated in the wrapped figure.
Additionally, object slippage occurs in 20\% of cases due to unstable contact, and in 16\% of cases, the robot exceeds torque limits by squeezing the object too hard. To address these failure scenarios, generating demonstrations from more diverse initial states, including near-failure situations, would be a promising avenue for future work.

\vspace{-0.5em}
\section{Conclusion}
\vspace{-0.25em}

We presented \MethodAcronym, a planning-guided diffusion policy learning method for generalizable contact-rich bimanual manipulation. Leveraging the recent advances in efficient model-based planning through contact, we generate large-scale, high-quality demonstration trajectories in simulation. We then perform visuomotor imitation learning using a task-conditioned point cloud diffusion policy, and we propose essential design choices in feature
extraction, task representation, and action prediction that enable effective policy generalization to unseen scenarios and sim-to-real transfer. Evaluations in both simulation and the real-world demonstrate our policy's effectiveness in manipulating objects with diverse geometries, dimensions, and physical properties.
In this paper, we have only used primitive shapes for data synthesis and demonstrated table-top manipulation tasks.
Future directions include further diversifying training environments with objects from large datasets and scaling up our approach to more dexterous and dynamic tasks.

\bibliographystyle{IEEEtran}
{\footnotesize
\bibliography{references.bib}}

\end{document}